\documentclass[fleqn]{article}
\usepackage{graphicx}
\usepackage{amssymb,amsmath,array}
\usepackage[margin=2.75cm]{geometry}

\usepackage{url} 
\usepackage{xspace}
\usepackage{subfigure}
\usepackage{hyperref}
\usepackage{bbm}

\newcommand{\kl}[2]{\textrm{KL}\!\br{#1|#2}}
\newcommand{\sett}[1]{\myset{#1}}
\newcommand{\myset}[1]{\mathcal{\uppercase{#1}}}
\newcommand{\const}{\textsf{const.}}

\newcommand{\wrt}{{\em w.r.t. \xspace}}

\newcommand{\ie}{{\em i.e. \xspace}}
\newcommand{\ocm}{\hspace{1cm}}
\newcommand{\hcm}{\hspace{0.25cm}}

\newcommand{\pdif}[2]{\frac{\partial}{\partial{#2}}#1}

\newcommand{\abs}[1]{|{#1}|}
\newcommand{\ind}[1]{\mathbb{I}\sq{#1}}

\newcommand{\dataind}{n}

\newcommand{\Dataind}{N}
\newcommand{\datadim}{D}

\newcommand{\bmp}{\begin{minipage}}
\newcommand{\emp}{\end{minipage}}
\newcommand{\beq}{\begin{equation}}
\newcommand{\eeq}{\end{equation}}

\newcommand{\tw}{\textwidth}

\newcommand{\trace}[1]{{\rm trace}\left({#1}\right)}
\newcommand{\pdist}{p}

\newcommand{\m}[1]{#1}
\renewcommand{\v}[1]{#1}
\newcommand{\ndist}[3]{{\cal{N}}\br{#1\thinspace\vline\thinspace #2,#3}}
\newcommand{\av}[1]{\left\langle{#1}\right\rangle}

\newcommand{\trans}{^{\textsf{T}}}
\newcommand{\cb}[1]{\left\{ {#1} \right\}}
\newcommand{\br}[1]{\left( {#1} \right)}
\newcommand{\sq}[1]{\left[ {#1} \right]}

\newcommand{\Id}{\m{I}}

\newcommand{\figref}[1]{Figure (\ref{#1})}
\renewcommand{\eqref}[1]{Equation(\ref{#1})}

\newcommand{\tabref}[1]{table(\ref{#1})}

\newcommand{\secref}[1]{section(\ref{#1})}

\newtheorem{mytheorem}{Theorem}

\newenvironment{proof}[1][Proof]{\begin{trivlist}
\item[\hskip \labelsep {\bfseries #1}]}{\end{trivlist}}
\newenvironment{definition}[1][Definition]{\begin{trivlist}
\item[\hskip \labelsep {\bfseries #1}]}{\end{trivlist}}
 
\begin{document}

\title{Variational Optimization}

\author{Joe Staines and David Barber\vspace{0.5cm}\\
University College London - Computer Science \\
Gower Street, London, WC1E 6BT - United Kingdom
}

\maketitle

\begin{abstract}
We discuss a general technique that can be used to form a differentiable bound on the optima of non-differentiable or discrete objective functions.  We form a unified description of these methods and consider under which circumstances the bound is concave.   In particular we consider two concrete applications of the method, namely sparse learning and support vector classification. 
\end{abstract}

\section{Optimization by Variational Bounding}

\label{variational}

We consider the general problem of function maximization, $\displaystyle\max_{x}f({x})$ for vector $x$. When $f$ is differentiable and $x$ continuous, optimization methods that use gradient information are typically preferred over non-gradient based approaches since they are able to take advantage of a locally optimal direction in which to search. However, in the case that $f$ is not differentiable or $x$ is discrete, gradient based approaches are not directly applicable. In that case, alternatives such as relaxation, coordinate-wise optimization and stochastic approaches are popular \cite{Lemarechal1989}. Our interest is to discuss another general class of methods that yield differentiable surrogate objectives for discrete $x$ or non-differentiable $f$.  The Variational Optimization (VO) approach is based on the simple bound\footnote{Note that this provides a bound on the optimum of $f(x)$, not necessarily on $f(x)$ itself.}
\begin{equation}
f^* = \max_{x \in \sett{C}}  f({x})  \geq  \big< f(x)\big>_{p({x}|\theta)} \equiv E(\theta)
\end{equation}
where $\av{\cdot}_p$ denotes expectation with respect to the distribution $p$ defined over the solution space $\sett{C}$. The parameters $\theta$ of the distribution $p(x|\theta)$ can then be adjusted to maximize the lower bound $E(\theta)$. This bound can be trivially made tight provided the distribution $p(x|\theta)$ is flexible enough to allow all its mass to be placed in the optimal state ${x}^*=\arg\max_{x} f(x)$.  

Under mild restrictions, the bound is differentiable, see \secref{app:diff}, and the bound is a smooth alternative objective function (see also \secref{sec:smoothing} on the relation to `smoothing' methods).  The degree of smoothness  (and the deviation from the original objective) increases as the dispersion of the variational distribution increases.  In \secref{app:theorem} we give sufficient conditions for the variational bound to be concave.  The purpose of this paper is to demonstrate the ease with which VO can be applied and to discuss its merits as a general way to construct a smooth alternative objective. 

\subsection{Differentiability of the variational objective}

\label{app:diff}
When $f(x)$ is not differentiable, under weak conditions $E(\theta)$ can be made differentiable.  The gradient of $E(\theta)$ is given by
\begin{equation}
\frac{\partial E}{\partial \theta} = \frac{\partial}{\partial \theta} \int_{\sett{C}} f(x)   p(x| \theta) dx.
\end{equation}
We can bring the differential under the integral sign
provided: 

(i)  $f({x})   p(x| \theta)$ is Lebesgue integrable and differentiable with respect to $\theta$

(ii) there exists an integrable function $F:\sett{C} \rightarrow \mathbbm{R}$ such that, for all $\theta$,
\begin{equation}
\left| \frac{\partial}{\partial \theta} f({x})   p(x| \theta) \right| < F(x)
\end{equation}  

These weak conditions mean that for a large-class of problems in which $f$ is non-differentiable or $x$ discrete, the bound $E(\theta)$ is differentiable with respect to the parameters $\theta$. For example, consider the non-differentiable objective $f(x)\equiv \ind{x\geq 0}$ with $x$ normally distributed with mean $0$ and unit variance, $p(x|\theta)=\ndist{x}{\theta}{1}$. Then $\frac{\partial E}{\partial \theta}=\exp(- \theta^2/2)/\sqrt{2\pi}$.

\subsection{Concavity of the variational bound}
\label{app:theorem}

Here we consider the special case of continuous $x$ and non-differentiable $f(x)$ and describe conditions which are sufficient for $E(\theta)$ to be concave in $\theta$\footnote{The variational objective $E(\theta)=\langle f(x)\rangle_{q(x|\theta)}$ is bounded above and below by the maximum and minimum of $f(x)$. Where the minimum is finite then, $E(\theta)$ cannot be strictly concave over all parameter space $\theta \in \mathbb{R}^N$. This prevents us from forming strictly convex problems for fully unconstrained $\theta$ whenever $f$ has finite minimum. However, even in this case, the bound may be quasi-concave (uni-modal).}. To do this we take advantage of a recent result by \cite{challis-barber-10} to prove the concavity of a Kullback-Leibler based variational approximation for generalized linear models. We first introduce the general concept of an expectation affine distribution.

\begin{definition}
\emph{(Expectation affine)}
A distribution $\pdist(x|\theta)$ is expectation affine if, for linear functions $\alpha$, $\beta$, distribution $q(z)$ and function $f$,
\beq
\big<f(x)\big>_{\pdist(x|\theta)}=\Big<f\big(\alpha(\theta) z + \beta(\theta)\big)\Big>_{q(z)}
\eeq
\end{definition}

\begin{mytheorem}
\label{th:main}
Let $f(x)$ be a concave function and $\pdist(x|\theta)$ an expectation affine distribution. Then $E(\theta)\equiv \av{f(x)}_{\pdist(x|\theta)}$ is concave in $\theta$.
\end{mytheorem}
\newcommand{\tlambda}{\tilde{\lambda}}
\begin{proof}
Defining $\tlambda\equiv 1-\lambda$ and using the fact that $p$ is expectation affine,
\begin{align}
E(\lambda\theta_1+\tlambda\theta_2)&  = \Big<f\big(\alpha(\lambda\theta_1+\tlambda\theta_2)z+
\beta (\lambda\theta_1+\tlambda\theta_2)\big)\Big>_{q(z)}\\
&=\bigg<f \Big(\lambda\big(\alpha \br{\theta_1}z+\beta\theta_1\big)+
\tlambda\big(\alpha\br{\theta_2}z+\beta(\theta_2)\big)\Big)\bigg>_{q(z)}
\end{align}
Since $f$ is concave, then
\beq
f(\lambda x_1+\tlambda x_2)\geq \lambda f(x_1)+\tlambda f(x_2)
\eeq
Hence
\begin{align}
E(\lambda\theta_1+\tlambda\theta_2) &\geq \lambda \Big<f\big(\alpha (\theta_1) z+\beta(\theta_1)\big)\Big>_{q(z)} +\tlambda\Big<f\big(\alpha(\theta_2)z+\beta(\theta_2)\big)\Big>_{q(z)}\\
&=\lambda E(\theta_1)+\tlambda E(\theta_2) 
\end{align}

\end{proof}
\subsubsection{Concavity for Gaussian $p$}
As an example, consider a multivariate Gaussian distribution $\pdist(\v{x}|\mu,\m{C})$ with mean $\mu$ and covariance $\Sigma=\m{C}\m{C}\trans$, where $\m{C}$ is the Cholesky factor. Then using the change of variables $\v{z}=\m{C}^{-1}\br{\v{x}-\mu}$ we have
\beq
\big<f(\v{x})\big>_{p\br{\v{x}|\mu,\m{C}}}=\big<f(\m{C}\v{z}+\mu)\big>_{\ndist{\v{z}}{\v{0}}{\Id}}
\eeq
That is, we can replace the average \wrt $\ndist{x}{\mu}{\Sigma}$ by an average with respect to the standard multivariate normal $\ndist{z}{0}{I}$ by using a linear transformation of parameters. To show that this is expectation affine, for $D$-dimensional vector $\v{x}$ and Gaussian $\pdist(\v{x}|\mu,\m{C})$ we define the $D+D(D+1)/2$  dimensional parameter vector $\theta$ which stacks the mean $\mu$ and lower triangular elements of $\m{C}$.  By defining $\alpha$ and $\beta$ as linear projections that select the $\m{C}$ and $\mu$ components respectively of $\theta$, we see that the multivariate Gaussian is expectation affine.  Hence, provided $f(x)$ is concave, then $E(\theta)=\big<f(x)\big>_{p(x|\mu,C)}$ is jointly-concave in $(\mu,C)$.

\begin{figure*}[t]
\begin{center}
\subfigure[]{\includegraphics[width=0.3\tw]{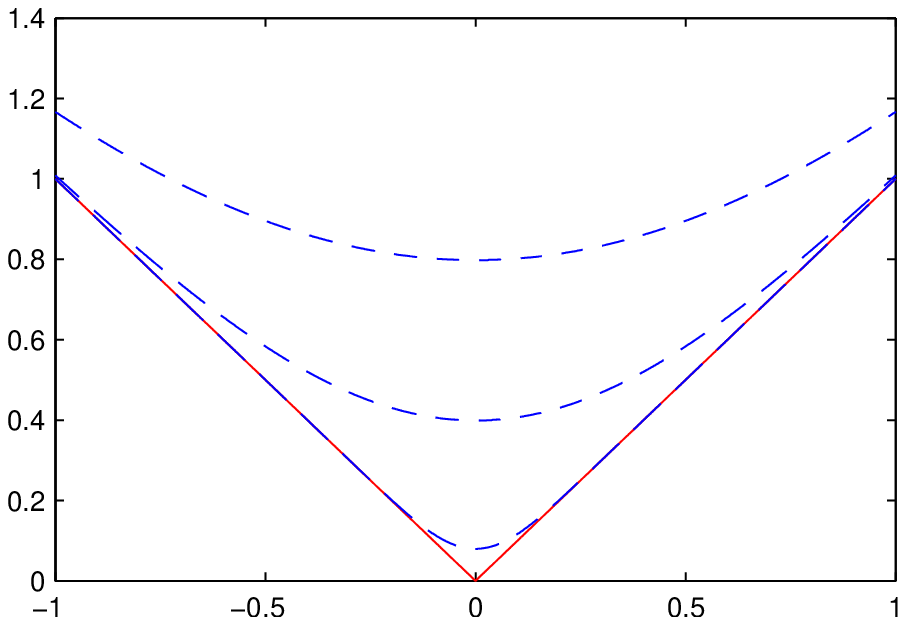}}
\subfigure[]{\includegraphics[width=0.3\tw]{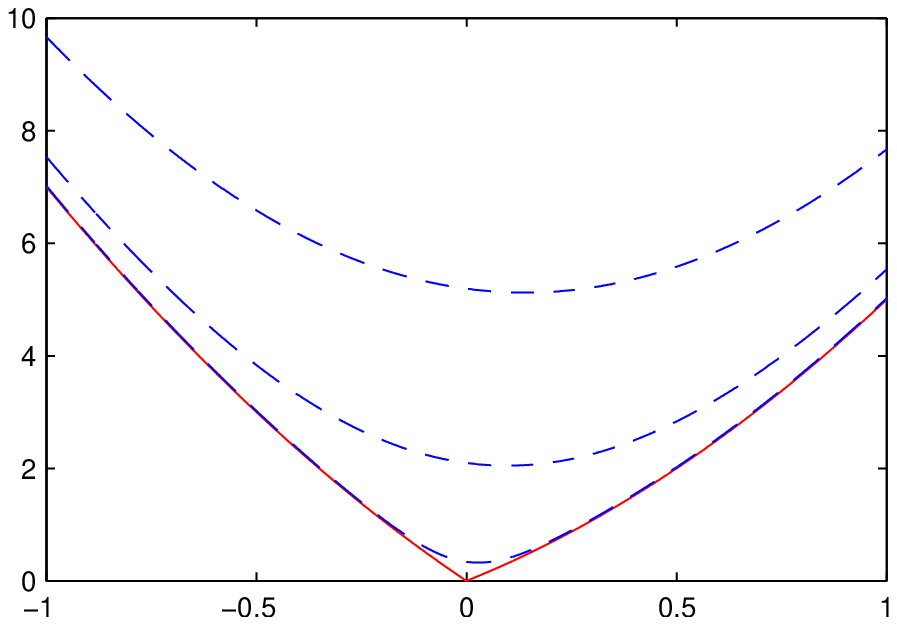}}
\subfigure[]{\includegraphics[width=0.3\tw]{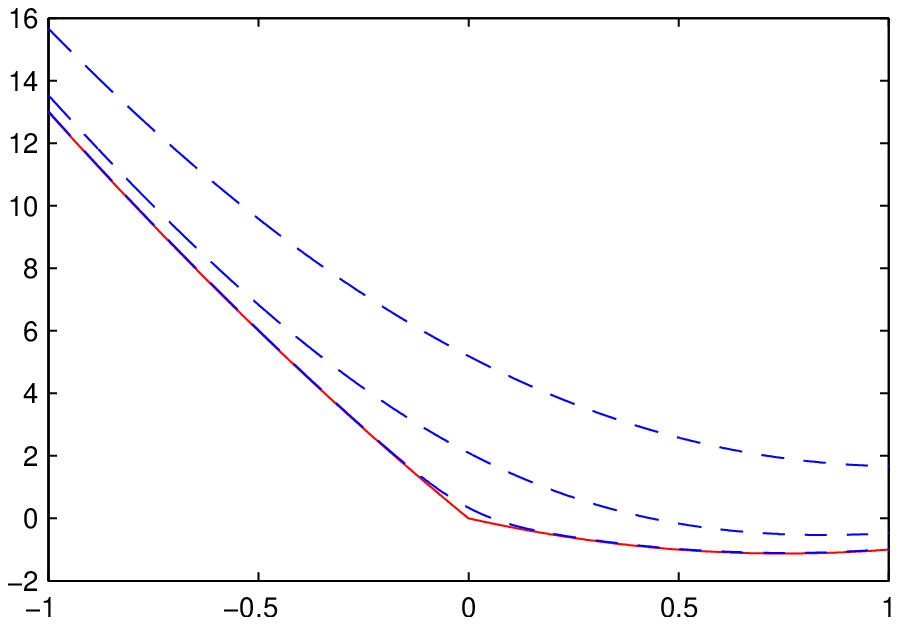}}
\end{center}
\caption{(a) The solid line represents the function $|w|$ and the dashed lines the bound $\av{|w|}_{\ndist{w}{\mu}{\sigma^2}}$ plotted as a function of $\mu$ for three different values $\sigma\in\cb{0.1,0.5,1}$ (from bottom to top in the figure).  (b) The lasso objective function $f(w)$ plotted for $A=2$, $\lambda=4$, $b=-1$, $c=0$. In this case the optimal setting for $w$ based on the lasso objective is $w=0$. However, the optimal setting that minimizes the upper bound is slightly larger than zero. This means that the bound does not force the weights to exactly zero. (c) The lasso objective for $A=2$, $\lambda=4$, $b=-7$, $c=0$. In this case the optimal setting for $w$ for both the lasso objective and the variational bound are non-zero and numerically very close. This is to be expected since, as we move further away from the origin, the variational approximation more closely matches the objective, resulting in a closer match of their optima.
}
\label{fig:L1}
\end{figure*} 

\subsection{Bound optimization}

VO provides a differentiable bound, whose optimum will still need to be found numerically. Since the bound if differentiable, any standard gradient-based method can be applied, such as Newton or Quasi-Newton methods. In the following sections, we discuss two example applications, the lasso and SVM objectives, both of which result in convex bound minimization problems. For the lasso problem we chose to use a simple diagonal approximation to Newton's method to optimize the bound since this is simple to implement and often surprisingly effective. Naturally, more sophisticated procedures could be applied, though this is not the main focus of our interest here.  For the SVM we chose L-BFGS, though again there are many alternatives that could be considered. 

\section{Sparse Least Squares Regression}
The lasso is a well-known least squares regression estimator that encourages sparse solutions through a one-norm regularizer \cite{tibshirani96regression}. For $D$-dimensional inputs $x^\dataind$, outputs $y^\dataind$, $\dataind=1,\ldots, \Dataind$, and positive regularising constant $\lambda$, the lasso objective is to minimize
\beq
f(w) \equiv \sum_{\dataind=1}^\Dataind \br{y^\dataind - w\trans x^\dataind}^2 + \lambda \sum_{i=1}^D |w_i| = c + w\trans b + w\trans A w + \lambda\sum_{i=1}^D|w_i|
\eeq
with
\begin{equation}
c\equiv \sum_\dataind \br{y^\dataind}^2, ~~~ b\equiv -2\sum_\dataind y^\dataind x^\dataind, ~~~ \text{and} ~~~ A\equiv \sum_\dataind x^\dataind {x^\dataind}\trans.
\end{equation}
The one-norm term $\sum_i |w_i|$ is non-differentiable at the origin and hence standard gradient based optimization algorithms cannot be directly applied.  Section \ref{app:theorem} shows that the variational bound conserves the convexity of an objective if a Gaussian (or more generally, expectation affine) distribution is used. Since the lasso objective is convex in $w$ it is natural to consider a Gaussian variational distribution $p(w|\theta)=\ndist{w}{\mu}{\Sigma}$ with mean $\mu$ and covariance $\Sigma$. This gives the upper bound $f^*\leq E(\mu,C)$, where
\begin{equation}
E(\mu,C) \equiv \sum _{\dataind} \av{\br{y^\dataind - w\trans {x}^\dataind}^2}_{\ndist{w}{\mu}{\Sigma}}
+ \lambda \sum_i \av{|w_i|}_{\ndist{w_i}{\mu_i}{\Sigma_{ii}}}.
\end{equation}
From Theorem \ref{th:main}, this is jointly convex in $(\mu,C)$ where $\Sigma=CC\trans$, the Cholesky decomposition. It is straightforward to show that
\beq
\av{\abs{w}}_{\ndist{w}{\mu}{\sigma^2}}=\mu \bigg(1-2\phi\Big(-\frac{\mu}{\sigma}\Big)\bigg)+2\frac{\sigma}{\sqrt{2\pi}} e^{-\frac{\mu^2}{2\sigma^2}}
\eeq
where $\phi(x)=\int_{-\infty}^x e^{-y^2/2}/\sqrt{2\pi}dy$ is the cumulative density function of the standard normal distribution. More explicitly,
\begin{equation}
E(\mu,C) = c + \mu\trans b + \mu\trans A \mu + \trace{A\Sigma} 
+ \lambda \sum_i \br{\mu_i\bigg(1- 2\phi\Big(-\frac{\mu_i}{\sigma_i}\Big)\bigg)+2\frac{\sigma_i}{\sqrt{2\pi}}e^{-\frac{\mu_i^2}{2\sigma_i^2}}}
\end{equation}
where $\sigma_i^2\equiv \Sigma_{ii}$.  The gradient and Hessian w.r.t $\mu$ are straightforward to compute using the results
\beq
\pdif{\av{\abs{w}}_{\ndist{w}{\mu}{\sigma^2}}}{\mu}=1-2\phi\br{-\frac{\mu}{\sigma}}, \hcm
\frac{\partial^2 E}{\partial \mu_i\partial \mu_j} = 2 \displaystyle A_{ij}  + \frac{2\lambda}{\sqrt{2\pi\sigma^2_i}} e^{-\frac{\mu_i^2}{2\sigma_{i}^2}} \delta_{ij}.
\eeq
Since the bound is non-differentiable for $\Sigma_{ii}=0$, we use a fixed isotropic covariance $\Sigma=\sigma^2\Id$ with $\sigma>0$ and minimize the bound with respect to $\mu$ alone.

\begin{figure}[t]
\bmp{0.3\tw}
\begin{center}
\includegraphics[scale=0.6]{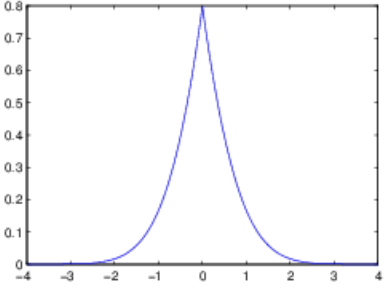}
\end{center}
\emp
\bmp{0.7\tw}
\caption{The function $z\big(1- 2\phi\br{-z}\big)+\sqrt{\frac{2}{\pi}}e^{-\frac{z^2}{2}}-\abs{z}$.\label{fig:EF}}
\emp
\end{figure}

\subsection{Bounding the error}
The difference between the bound and the function is given by (see \figref{fig:L1})
\begin{equation}
E(\mu) - f(\mu) = \trace{A\Sigma}
+\lambda\sum_i\sigma_i\br{z_i\br{1- 2\phi\br{-z_i}}+\sqrt{\frac{2}{\pi}}e^{-\frac{z_i^2}{2}}-\abs{z_i}}
\end{equation}
where $z_i\equiv\mu_i/\sigma_i$. This splits into a set of independent terms, each of which has its maximum at $z_i=0$, \ie $\mu_i=0$, giving a maximal error of $\trace{A\Sigma}+\frac{2\lambda}{\sqrt{2\pi}}\sum_i \sigma_i$.
Hence to guarantee to find a surrogate objective whose optimum value is within $\Delta_f$ of the true lasso optimal value we must use, for $\Sigma=\sigma^2\Id$,
\beq
\sigma\leq \frac{1}{\sqrt{\trace{A}}}\br{\sqrt{\br{\frac{\lambda^2D^2}{2\pi}+\Delta_f\trace{A}}}-\frac{\lambda D}{\sqrt{2\pi}}}.
\eeq
Hence, provided that $\sigma$ is set small enough and assuming that the optimum of the bound is found accurately, we can guarantee to find the optimum value of the lasso objective to within some specified tolerance.

\subsection{Lasso Experiments}
\label{lassoexpt}

We first generated a sparse $D$-dimensional parameter vector with components $u_i$ sampled according to
\beq
u_i\sim \left\{ \begin{array}{ll}
                  0 & \text{with probability 0.5} \\
                  \ndist{u_i}{5}{1} & \text{with probability 0.25}\\
                  \ndist{u_i}{-5}{1} & \text{with probability 0.25}\\
                \end{array}
\right.
\eeq
We then created a set of $\Dataind=10D$ training points in $D$-dimensional space from the standard multivariate normal distribution $\ndist{x}{0}{\Id}$. The clean labels for these points were given by $y^\dataind_0=u\trans x^\dataind$ and the final noisy labels by $y^\dataind=y^\dataind_0+\epsilon^\dataind$ where $\epsilon^\dataind$ is Gaussian random noise with mean zero and standard deviation $0.1 \frac{1}{\Dataind} \sum_{\dataind=1}^\Dataind |y_0^\dataind|$, chosen to obscure but not dominate the clean labels. We set $\lambda=30D$ to give solutions with sparsity roughly the same as the initial parameter vector $u$.

At each iteration we fixed $\Sigma$ and updated $\mu$ using a diagonal approximation to the Newton method (to avoid the cost of inverting the full Hessian). For gradient $E'_i$ and Hessian elements $E_{ii}''$ the updates were 
\beq
\mu^{new}_i = \mu_i^{old} -0.1 \frac{E'_i}{E''_{ii}}.
\eeq 
There is a trade-off between using a large $\sigma$, resulting in a bound whose optimum can be numerically obtained easily versus using a small $\sigma$ for which the optimum of the bound is more difficult to find numerically, but for which the resulting bound optimum more closely matches the optimum of the true objective. Our experience is that in practice it is therefore best to start iterative procedures that optimize the bound with a relatively large $\sigma$, reducing this with each iteration. For this particular set of experiments we used an initial covariance matrix of 0.1 times the identity and reduced it by a factor of 0.9 at each iteration. The initial estimate for $\mu$ was a vector of zeros. We terminated when the mean absolute difference in the solution elements between iterations was less than $10^{-15}$ of the mean absolute value of the elements of the current solution and took this terminal $\mu$ to be the estimate for $\arg\min_{w}{f(w)}$.

\begin{table}[t]
\centering
\scalebox{1.0}{
\small
\begin{tabular}{|l| c | c | c |}
\multicolumn{2}{c}{} & \multicolumn{2}{c}{Small problems - $D=50$} \\
\cline{1-1} \cline{3-4}
\rule{0cm}{0.4cm} Algorithm & & \multicolumn{1}{|l|}{Solution time (SD)}& \multicolumn{1}{|l|}{Relative Error (SD)} \\
\cline{1-1} \cline{3-4}
Variational optimization \rule{0cm}{0.35cm}  && 0.1269  (0.0680) & $ 3.791 \times 10^{-14}~ (5.62\times 10^{-14}) $ \\
Shooting && 0.0142 (0.0037) & $ 2.485 \times 10^{-16}~ (2.21\times 10^{-16}) $ \\
Iterated ridge regression && 0.0268 (0.0208)  & $ 1.851 \times 10^{-10}~ (6.27\times 10^{-10}) $ \\
Smoothing (integral sigmoid) & &  0.1088 (0.0603) & $ 4.421 \times 10^{-15}~ (3.06\times 10^{-15}) $ \\
Smoothing ($\sqrt{x^2 + \epsilon}$) & & 0.0830 (0.0178)& $ 4.916 \times 10^{-15}~ (1.62\times 10^{-14}) $ \\
Projection && 0.0224 (0.0107) & $ 2.672 \times 10^{-16}~ (1.80\times 10^{-15}) $ \\
Sub-gradients && 0.0472 (0.0180) & $ 9.176 \times 10^{-10}~ (8.31\times 10^{-09}) $\\
\cline{1-1} \cline{3-4}
\multicolumn{2}{c}{} & \multicolumn{2}{c}{\rule{0cm}{0.6cm} Larger problems - $D=200$} \\
\cline{1-1} \cline{3-4} 
\rule{0cm}{0.4cm} Algorithm && \multicolumn{1}{|l|}{Solution time (SD)}& \multicolumn{1}{|l|}{Relative Error (SD)} \\
\cline{1-1} \cline{3-4} 
Variational optimization \rule{0cm}{0.35cm} && 0.1308 (0.0405) & $1.923\times 10^{-15}~ ( 6.01\times 10^{-16})$ \\
Shooting && 0.0676 (0.0209) & $2.656\times 10^{-16} ~( 3.17\times 10^{-16})$  \\
Iterated ridge regression && 0.1645 (0.0844)  & $4.004\times 10^{-14}~ (1.19\times 10^{-13}) $ \\
Smoothing (integral sigmoid) & &  1.2305 (0.3381) & $2.638\times 10^{-15}~ (9.19\times 10^{-16})$ \\
Smoothing ($\sqrt{x^2 + \epsilon}$) & & 1.1385 (0.3673)& $1.471\times 10^{-15}~ (1.86\times 10^{-15})$ \\
Projection && 0.3121 (0.0726) & $5.325\times 10^{-16}~ (2.85\times 10^{-16})$ \\
Sub-gradients && 1.3478 (0.3273) & $ 6.038 \times 10^{-11}~ (9.42\times 10^{-10}) $\\
\cline{1-1} \cline{3-4}                                
\end{tabular}}
\caption{Performance and time taken (in seconds) for algorithms solving lasso problems of two different sizes. All algorithms were implemented in MATLAB and run on a 2.27GHz 4GB machine. The averages and the standard deviations result from 500 experiments.}
\label{tab:lasso}
\end{table}
We tested our method against a number of standard lasso solvers in Mark Schmidt's {\tt{minFunc}} MATLAB package\footnote{{\tt{www.di.ens.fr/$\sim$mschmidt/Software/minFunc.html}}} to gain an indication of its performance relative to established approaches. Since the true optima are not known, we measured the success of the algorithms relative to the best solution $f_{\text{best}}$, found by any of the algorithms on each problem instance. In Table \ref{tab:lasso} we give the mean and standard deviation of relative errors $(f-f_{\text{best}})/(f_{\text{best}})$.  The results indicate that VO is capable of approximating the global optimum in moderate time and that it scales well with problem size. In particular, VO provides solutions of similar quality to other smoothed methods in \cite{mschmidt} (the first based on approximating the $L_1$ norm using an integral of two sigmoid functions and the second using $\sqrt{x^2+\epsilon}$ where $\epsilon$ is some positive constant)\footnote{The improvement of VO over other smoothing approaches is most likely accounted for by implementation differences and our reduction of $\sigma$ at each iteration.}.

\subsection{Fused Lasso Sparse Regression}

From \tabref{tab:lasso}, the standard lasso problem is most effectively solved by the shooting algorithm \cite{fu1998}.  
Shooting corresponds to solving each component whilst keeping the others fixed, and subsequently cycling through components to convergence. This is particularly effective in the standard lasso problem since the corresponding one-dimensional optima for each component can be found in closed form; also the objective only weakly couples the components of the vector $w$.  In contrast, the fused lasso problem induces additional sparsity between adjacent elements using the objective:
\beq
f(w)= c + w\trans b + w\trans A w + \lambda_1 \sum_{i=1}^{\datadim}|w_i|+ \lambda_2 \sum_{i=2}^{\datadim}|w_i-w_{i-1}|
\eeq
The additional term also introduces strong dependencies between adjacent components and componentwise ('shooting') methods struggle to converge \cite{friedman-etal-07}. In applying VO, with  $\Sigma=\sigma^2\Id$, the additional terms $\av{ \abs{ w_i- w_{i-1}}}_{\ndist{w}{\mu}{\sigma^2\Id}}$ in the bound are given by:
\begin{align}
(\mu_i-\mu_{i-1}) \bigg(1-2\phi\Big(\frac{\mu_{i-1}-\mu_{i}}{\sqrt{2}\sigma}\Big)\bigg)+\frac{2\sigma}{\sqrt{\pi}} e^{\frac{-(\mu_{i}-\mu_{i-1})^2}{4\sigma^2}}
\end{align}
As for the standard lasso case, we can readily obtain the gradient and Hessian of the objective.

\subsection{Fused Lasso Experiments}

We set the $D$-dimensional parameter vector so that there would be some real underlying sparsity in the differences between elements of the optimum $w$. The first element in the parameter vector $u_1$ was generated as in the previous problem, then all subsequent elements were generated by:
\beq
u_{i>1}\sim \left\{ \begin{array}{ll}
						u_{i-1} & \text{with probability 0.5}\\
                  0 & \text{with probability 0.25} \\
                  \ndist{u_i}{5}{1} & \text{with probability 0.125}\\
                  \ndist{u_i}{-5}{1} & \text{with probability 0.125}\\
                \end{array}
\right.
\eeq
We again created a set of $\Dataind=10D$ training points in $D$-dimensional space from the standard multivariate normal distribution. The clean labels for these points were then given by $y^\dataind_0=u\trans x^\dataind$ and the final noisy labels by $y^\dataind=y^\dataind_0+\epsilon^\dataind$ where $\epsilon^\dataind$ is random noise with mean zero and standard deviation $0.1 \frac{1}{\Dataind} \sum_{\dataind=1}^\Dataind |y_0^\dataind|$.

We ran a $D=500$ experiment 1000 times with $\lambda_1=500$ and $\lambda_2=200$. These values were chosen to give rise to a problem in which the loss function had a significant impact but did not enforce more sparsity than was typically present in the original parameter vector $u$. The initial standard deviation was 0.1, and we shrunk it by 0.9 at each iteration. For comparison we used the SLEP package \cite{liu2010}, which has very competitive performance compared to other state-of-the-art solvers. The SLEP package is based on a version of Nesterov's method \cite{nesterov}, a two step gradient method with backtracking line search. Using a shrinking variance and a convergence tolerance of $10^{-6}$, the mean relative error in terms of function value of VO compared to SLEP was $1.59\times 10^{-4}$. The mean of the relative distances $L_2(x-y)/L_2(y)$ between each VO solution $x$ and SLEP solution $y$ was 0.0135. The mean CPU time for VO was 0.0947s and for the SLEP algorithm 0.0724s\footnote{SLEP is implemented in C, so the comparison of speed with our MATLAB implementation is not definitive.}. Compared to other approaches, our method is very simple to implement and has good performance compared to the state of the art.

\section{Soft Margin Support Vector Machines}
The support vector machine (SVM) is a method for finding the optimal separating hyperplane of set of $N$ data points $x^n$ with labels $y^n=\pm 1$. This separating hyperplane may be in the space of the data points, or in a higher dimensional feature space if linear separation is not possible. In this second case, the data points are mapped to the feature space with a function $\varphi(x^n)$. When the data are not separable the soft margin support vector machine is used. This adds slack variables $\xi^n$ which allow misclassified data points at a cost controlled by a variable $C$. The complete soft margin support vector machine has primal form:  
\begin{align}
&\displaystyle\min_{w,b,\xi} w^{\mathsf{T}}w + C \displaystyle\sum_{n=1}^N \xi^n\\
&\text{subject to} ~~y^n(w^{\mathsf{T}}\varphi(x^n) + b ) \geq 1- \xi^nC ~~\text{and}~~\xi^n \geq 0.
\end{align}
This is classically solved by the method of Lagrange multipliers \cite{cortes1995}. This yields a convex, constrained quadratic program referred to as the dual problem. The dual form is often preferred because it provides a convenient way to deal with the constraints, and because kernel methods can be easily applied.

As an alternative, the primal problem can be re-expressed as a convex, unconstrained problem (see, for example, \cite{chapelle2007}). To remove the constraints from the primal problem, note that the objective is minimized when the slack variables $\xi^n$ are as small as possible without violating the constraints. This is achieved by setting $\xi^n=\max\{y^n(w^{\mathsf{T}}\varphi(x^n) + b)-1,0\}$, yielding the objective to be minimized:
\begin{equation}
f(w,b) = w^{\mathsf{T}}w + C \displaystyle\sum_{n=1}^N \max\{1-y^n(w^{\mathsf{T}}\varphi(x^n) + b),0\}.
\end{equation}
These slack variable terms are hinge loss penalties on misclassified data points. The hinge loss is convex in $w$ and $b$, thus the whole objective (a sum of hinge losses plus a convex quadratic term) is convex. It has discontinuities across each of the hyperplanes $w^{\mathsf{T}}\varphi(x^n) + b=y^n$.  To allow feature mapping we must solve the primal problem in a reproducing kernel Hilbert space without using Lagrange multipliers. Without loss of generality, we may assume that the solution $w$ can be expressed as
\beq
w= \sum_{n=1}^N \beta^n y^n \varphi(x^n)
\eeq
which gives rise to the kernelized objective function:
\beq
f(\beta ,b)=  \beta\trans K \beta + C \sum_{n=1}^N \max\{1-\sum_{m=1}^{N} K_{nm}\beta^m - b y^n ,0\}
\label{primalRKHS}
\eeq
where the kernel matrix is given by $K_{nm} = y^n y^m {\varphi(x^n)}^{\mathsf{T}}\varphi(x^m)$. This kernelized problem is convex; however it is discontinuous and therefore not directly amenable to gradient descent methods. 

The variational bound is the expectation of $f$ with respect to distributions over $\beta$ and $b$. We choose again Gaussian distributions over the parameters: $\beta$ distributed with mean $\mu_{\beta}$ and covariance $\sigma^2I$ and $b$ independently with mean $\mu_b$ and variance $\sigma^2$. This gives the bound $f^*\leq E$ with 
\begin{equation}
E\equiv \langle f(\beta,b)\rangle_{p(\beta,b| \mu_{\beta},\mu_b)} = \mu_\beta \trans K \mu_\beta + \text{trace}(\sigma^2K) + C \sum_{n=1}^N \left( \nu^n \phi \left(\frac{\nu^n}{\varsigma^n}\right) + \frac{\varsigma^n}{\sqrt{2 \pi}}e^{-\frac{1}{2}\left(\frac{\nu^n}{\varsigma^n}\right)^2} \right)
\end{equation} 
where
\beq
\nu^n=1- \sum_{m=1}^N K_{nm} \mu_{\beta}^m  + \mu_b y^n, \ocm \varsigma^n=\sqrt{\sigma^2+\sum_{m=1}^{N} K_{nm}^2 \sigma^2}.
\eeq
The gradients used in optimizing the variational objective can then also be calculated explicitly:
\begin{align}
\frac{\partial E }{\partial \mu_{\beta}^n} &= \sum_{m=1}^N 2 K_{nm} \mu_{\beta}^m - C \sum_{m=1}^N \frac{K_{nm}}{\varsigma^m}  \phi\left( \frac{\nu^m}{\varsigma^m} \right)\\
\frac{\partial E}{\partial b} &= - C \sum_{m=1}^N \frac{y^m}{\varsigma^m} \phi\left( \frac{\nu^m}{\varsigma^m} \right)
\end{align}

We can also bound the difference between the variational and original objectives, just as we did in the lasso case.
The difference between the bound and the function is given by
\begin{equation}
E(\mu_{\beta},\mu_{b}) - f(\mu_{\beta},\mu_{b}) = \trace{\sigma^2 K}
+ C \sum_{n=1}^N \left( \nu^n \phi \left(\frac{\nu^n}{\varsigma^n}\right) + \frac{\varsigma^n}{\sqrt{2 \pi}}e^{-\frac{1}{2}\left(\frac{\nu^n}{\varsigma^n}\right)^2}  -  \max\{\nu^n ,0\} \right)
\end{equation}
The terms of the sum are each maximized at $\nu^n=0$. $\nu^n$ are not independent, so these conditions are unlikely to occur simultaneously, but we can bound the difference by
\begin{equation}
E(\mu_{\beta},\mu_{b}) - f(\mu_{\beta},\mu_{b}) \leq \trace{\sigma^2 K} + C \sum_{n=1}^N \frac{\varsigma^n}{\sqrt{2 \pi}}.
\end{equation}
To guarantee to find a surrogate SVM objective whose optimum is within a specified function tolerance $\Delta_f$, one must thus satisfy
\beq
\sigma\leq \frac{1}{\sqrt{\trace{K}}}\br{\sqrt{\br{\frac{ C^2 M^2}{8\pi}+\Delta_f\trace{K}}}-\frac{C M}{\sqrt{8\pi}}}
\eeq
where
\beq
M=\displaystyle\sum_{n=1}^N \sqrt{1+\sum_{m=1}^N {K_{nm}^2}}.
\eeq

\begin{figure*}[t]
\begin{center}
\subfigure[]{\includegraphics[width=0.45\tw]{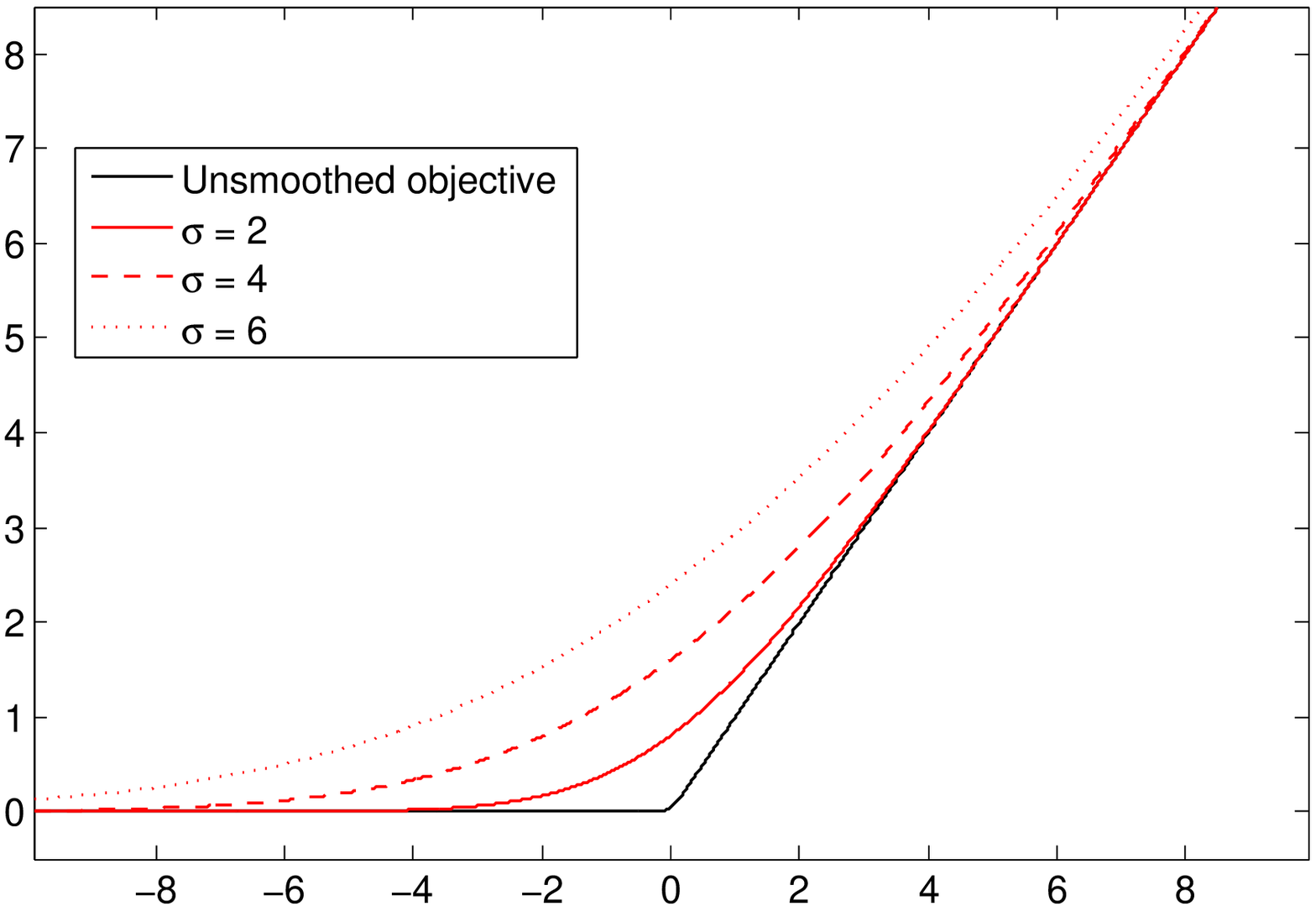}}\ocm
\subfigure[]{\includegraphics[width=0.45\tw]{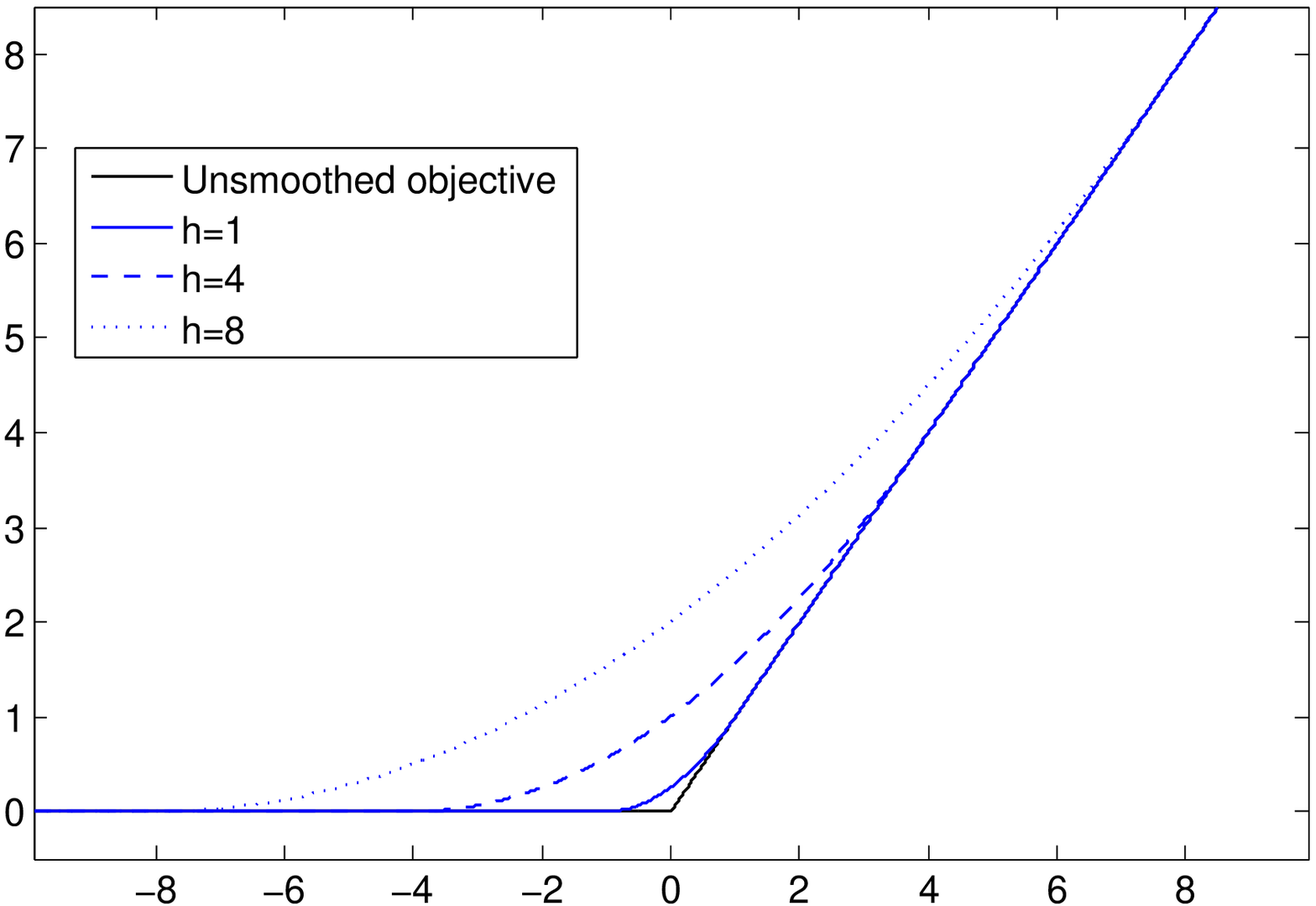}}\\
\subfigure[]{\includegraphics[width=0.45\tw]{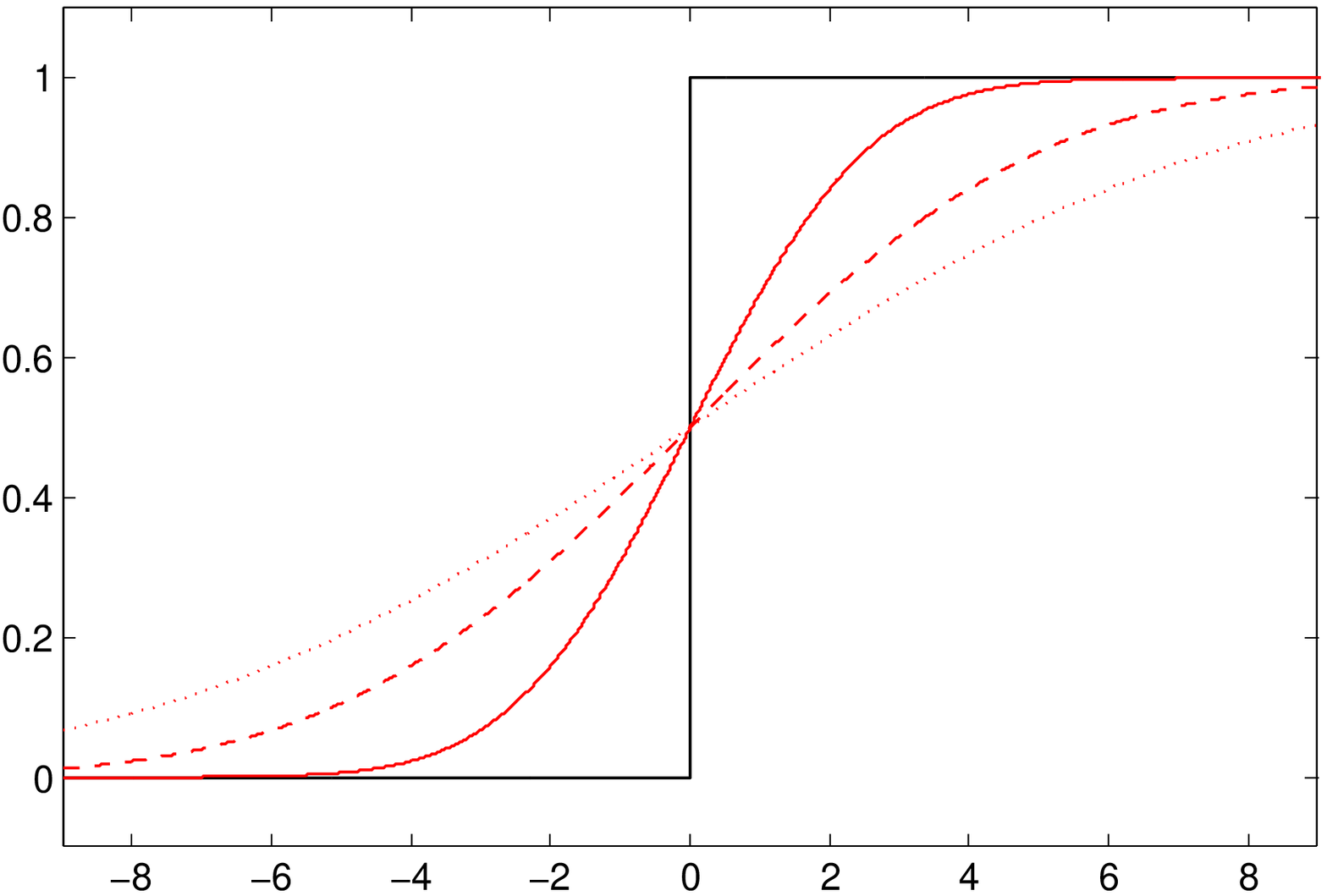}}\ocm
\subfigure[]{\includegraphics[width=0.45\tw]{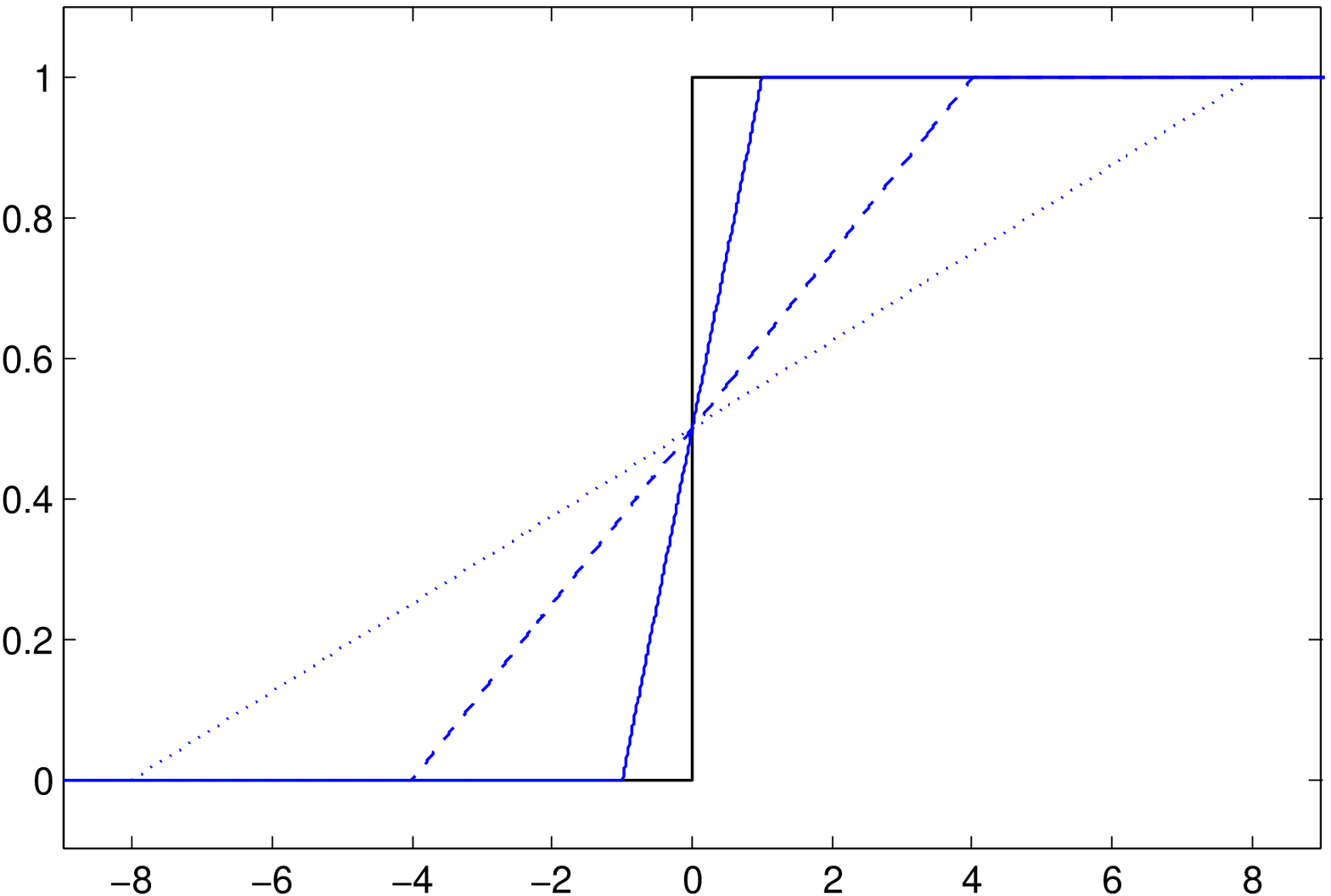}}
\end{center}
\caption{(a) hinge loss and VO smoothed hinge loss; (b) Huber loss; (c) smoothed hinge loss gradient; (d) Huber loss gradient.\label{fig:lascomp}}
\end{figure*}

\subsection{Chapelle's smoothing method}

Chapelle \cite{chapelle2007} formulates the problem as in \eqref{primalRKHS}, though allowing for any form of loss function $L$ rather than just the hinge loss.
\begin{equation}
E_{\text{Chapelle}}(\beta ,b)= \beta\trans K \beta + C \sum_{n=1}^N L(1-\sum_{m=1}^{N} \beta^m K_{nm} - b y^n )
\end{equation}
He then derives Newton methods with quadratic loss and, more relevant to our problem, the Huber loss:
\begin{equation}
L_{\text{Huber}}(\xi)  = \left\{ \begin{array}{ll}
\frac{(1+h \xi)^2}{4h} & \text{if} ~~| \xi|<h \\
\max\{\xi,0\} & \text{otherwise}\end{array}\right.
\end{equation}
The form of the Huber loss can be seen in \figref{fig:lascomp}. It is an upper bound on the hinge loss, $f^*\leq E_{\text{Chapelle}}$, and becomes tighter for smaller $h$. This bound can be used in the same way as the variational bound, giving a smoothed approximation to the SVM objective. As with VO, the objective is a sum of convex terms and is thus itself convex.  The Hessian of $E_{\text{Chapelle}}$  is given by
\begin{equation}
H= \left( \begin{array}{ll}
\frac{\partial^2 E_{\text{Chapelle}}}{\partial b^2} & \frac{\partial^2 E_{\text{Chapelle}}}{\partial b \partial \beta}\\
\frac{\partial^2 E_{\text{Chapelle}}}{\partial \beta \partial b} & \frac{\partial^2 E_{\text{Chapelle}}}{\partial \beta^2} \end{array} \right)
=2C \left( \begin{array}{ll}
1^{\mathsf{T}} I^{q} 1 & 1^{\mathsf{T}} I^{q} K \\
K I^{q} 1 & \frac{1}{C} K + K I^{q} K \end{array}\right)
\label{chaphess}
\end{equation}
where $1$ is a vector of ones of dimension $N$ and $I^{q}$ is a diagonal matrix indicating whether a given data point is in the quadratic portion of the loss function:
\begin{equation} 
I^{q}_{nn}=\left\{ \begin{array}{ll}
1 & \text{if} ~~ |1-\sum_{m=1}^N \beta^m K_{nm} + b y^n|<h \\
0 & \text{otherwise}\end{array}\right.
\end{equation}

Given that we wish to optimize the hinge loss, we must choose $h$ to be small. The quadratic region in the Huber loss is then small and as a consequence it is likely that for some trial solutions no training points will lie in this quadratic portion. This results in a value of zero for all Hessian elements corresponding to the offset (all blocks in the black matrix in  \eqref{chaphess} except the bottom right). The Hessian is therefore non-invertible and Chapelle's Newton update cannot be evaluated. If instead the L-BFGS method is used \cite{lbfgs} it can easily be shown that the approximate Hessian is always positive definite and thus Newton-like performance can be achieved without risking invalid updates. This also means the cubic cost of inversion in the full Newton method is avoided.

\subsection{SVM Experiments}
Synthetic problems were constructed by generating a random separation vector of length 3.5  and dimension 100. The length of this vector was chosen to give a problem which could be mostly, but not completely, separated using a linear kernel. The positively labelled points were then distributed with a standard multivariate normal about this separation vector and the negatively labelled points with the same distribution about the origin. The objective we used had a cost coefficient $C=10$. We define approximate convergence as finding a function value with relative error ($\frac{f^*-f}{f}$) of less than 0.1\% compared to the function value $f^*$ of the MATLAB quadratic program on the dual when run until completion. For each of 50 training sets for each training set size, each method was given the same initial solution (a vector generated from a standard normal distribution) and run until approximate convergence.

We compared VO to a modification of Chapelle's method (with and without shrinking the Huber parameter), sequential minimal optimization (SMO) and quadratic programming on the dual problem. We applied Chapelle's method by optimizing a Huber loss smoothed function using the L-BFGS implementation in the {\tt{minFunc}} package. To ensure comparability we treated VO in the same way, with the additional consideration of variance reduction at each iteration. We shrank the variance by a factor of 10 every 250 iterations, from an initial value of 0.001.  The Huber loss function was similarly shrunk by a factor of 10 every 250 iterations, but from an initial value of 10. SMO \cite{platt1999} solves the dual by analytically solving a sub-problem in two components and iterating until convergence. We used the implementation in the BRML toolbox \cite{brml}, which is based on the working set selection proposed by \cite{fan2005}. The dual was solved using the MATLAB built in function, {\tt{quadprog}}.

\begin{figure}[t]
\begin{center}
\includegraphics[width=0.8\tw]{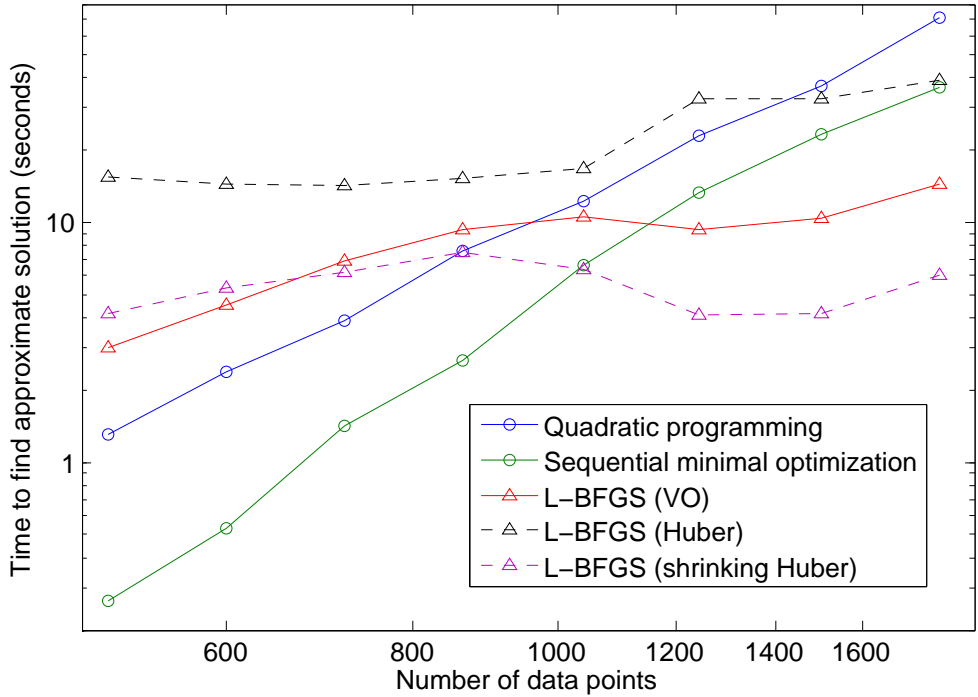}
\end{center}
\caption{A plot of the log of approximate solution time for SVM problems of varying size. For smaller problems SMO is the fastest method, but for larger ones it is beaten by variational optimization.}
\label{fig:svm}
\end{figure}

\figref{fig:svm} shows VO scaling better than both quadratic programming and SMO with respect to the number of training points. SMO is highly effective for smaller problems. This makes sense in the context of the findings of \cite{platt1999}; for sparse problems (since they have fewer support vectors) SMO is extremely fast. What is perhaps more surprising is that VO outperforms the Chapelle method (without shrinking) roughly by a constant factor. This outperformance doesn't come from faster evaluation of the gradient (quite the opposite; evaluation of a Huber loss gradient takes on average 0.0102s on our machine compared to 0.0237s for the gradient of the variational objective). Thus, we must conclude that there is some inherent advantage in the variance shrinking schedule which allows quicker convergence. 

The advantage of shrinking the variance (or Huber parameter) can be explained by imagining the impact of updates. \figref{fig:lascomp} shows the function value and gradient of both the Huber loss and VO smoothed loss. L-BFGS updates use an approximate Hessian, rather than the exact Hessian of a Newton update, but the behaviour is similar. These Newton type updates will fall further from optima when the Hessian varies more quickly. The figure makes it clear that the Hessian changes more quickly in the solution space when the variance is larger. Thus large variance helps to avoid slow convergence by minimizing overshooting of updates.

The impact of increasing the Huber parameter is similar, increasing the region about zero for which the gradient is constant. It would seem that so long as the initial value of $h$ is sufficiently large, the same benefit should be found as with VO; and indeed this is borne out by the results in \figref{fig:svm}. 

\section{Relation to Other Optimization Methods}
\label{sec:relation}

To our knowledge, relatively little attention has been given to the method of variational optimization and its relation to other approaches. In particular, the use of analytic averages to calculate gradients seems to be less considered.

In \cite{Berny-02,Berny-01} Berny considers a form of VO for general binary optimization $x_i\in\cb{0,1}$ through the use of a factorized distribution $p(x|\theta)=\prod_i \theta_i^{x_i}\br{1-\theta_i}^{1-x_i}$. In contrast to analytical VO, in Berny's work samples are drawn from $p(x|\theta)$ to approximate the expectations, rather than averages being computed analytically. Analytical bounds for binary problems are often easy to compute. Consider the binary quadratic program, in which one optimizes
\begin{equation}
f({x})={x}^\mathsf{T}{A}{x} + {b}^\mathsf{T}{x}
\end{equation}
where $x$ is constrained on the binary hypercube $x_i \in \{0,1\}$. The variational lower bound on the optimal solution is given by
\begin{align}
E(\theta) &\equiv  \av{{x}^\mathsf{T}{A}{x} + {b}^\mathsf{T}{x}}_{p(x|\theta)}\\
& =\sum_{i\neq j}A_{ij}\theta^{11}_{ij}+\sum_i \br{b_i+A_{ii}} \theta_i^1
\label{bqp}
\end{align}
One may then proceed to optimize $E(\theta)$, $\theta_i\in\sq{0,1}$, using any method of choice. However, we have not given this problem further consideration in this paper since it gives rise to non-convex problems and as such the quality of the solution is entirely determined by the specific optimization algorithm used. Note also that the binary quadratic programme is in general NP-hard, so that it would be unreasonable to expect any single algorithm to perform well for arbitrary $A$.  Whilst a numerical analysis of the different algorithms for various settings of $A$ would be interesting, it goes beyond the scope of the current paper. 

\subsection{Relation to smoothing\label{sec:smoothing}}

Smoothing methods (see for example \cite{Lemarechal1989}) replace an objective $f(x)$ by a `smoother' differentiable version $g(x)$, with the property that $g(x)$ will be easier to optimize than $f(x)$. Typically smoothing methods have a parameter that controls how to interpolate between $f(x)$ and smoother versions thereof.  Whilst VO also makes a smooth objective, it is not formally the same as a classical smoothing approach since, for the VO objective $E(\theta)$, $\theta$ inhabits a potentially different parameter space than $x$. 

\subsection{Relation to relaxation}
In solution set relaxation methods the constraint $x\in\sett{C}$ is replaced by a relaxed constraint $\sett{D}\supset\sett{C}$. Such relaxations provide an upper bound on $f^*$. That is
\beq
\av{f({x})}_{\pdist({x}|\theta)}\leq \max_{{x}\in\sett{C}} f({x})\leq  \max_{{x}\in\sett{D}} f({x})
\eeq
This means that we may bracket the optimum by the combined use of relaxation and variational optimization. Note that VO provides a lower bound on the optimum for all values of the variational parameter $\theta$, whilst the relaxation is not guaranteed to provide a bound for all $x\in\sett{D}$.

\subsection{Free energy approach}
Defining the distribution
\beq
\tilde{p}(x)=\frac{1}{Z}e^{\beta f(x)}, \hcm \beta\geq 0
\eeq
where $Z$ normalizes $\tilde{p}$, then we can find an approximation to $\tilde{p}(x)$ based on minimising
\beq
\kl{q}{\tilde{p}}\equiv \underbrace{\av{\log p(x|\theta)}_{p(x|\theta)}-\beta\av{f(x)}_{p(x|\theta)}}_{E(\theta)}+\const
\eeq
Typically the family of distributions $p(x|\theta)$ is chosen to ensure that $E(\theta)$ is tractably computable. For a fixed $\beta$, one then minimizes $E(\theta)$. Gradually, as $\beta$ is increased towards infinity, the distribution $p(x)$ becomes more sharply peaked around its maximal state, so that the approximating distribution $p(x|\theta)$ will also tend to this state. This method is quite general and applicable to either discrete or continuous optimization problems, see for example \cite{gallagher-frean-05}.

Whilst the method does not appear to provide a bound on $f^*$ in general, there are cases in which this approach does provide a bound. In particular, if the entropic term $\av{\log p(x|\theta)}_{p(x|\theta)}$ is constant with respect to $\theta$, then minimising $E(\theta)$ is equivalent to maximising $\av{f(x)}_{p(x|\theta)}$. In this case, the expectation is a bound on the optimum $f^*$. For example, for Gaussian $p(x|\theta)=\ndist{x}{\mu}{\Sigma}$ the entropy is a function of the covariance $\Sigma$ alone, so that for fixed covariance, free energy optimization is equivalent to optimizing the VO bound (using a fixed covariance and optimizing with respect to the mean).  In general, however, the two approaches are different.

\subsection{Estimation of Distribution Algorithms}

Estimation of distribution algorithms (EDAs) are a broad set of optimization algorithms for the problem $\max_w f(w)$. An EDA starts with a prior distribution $p_0(w)$ over the solution space. At each iteration this is then in used to generate a new set of solutions $\{w^n\}$. The function values $f(w^n)$ of these solutions are calculated and the distribution for the next iteration is constructed by some function (whose specification determines the type of EDA in question): $p_{t+1}(w)=F\bigl(p_{t}(w),\{f(w^n)\},\{w^n\}\bigr)$. If VO were used, with expectations $\langle f(w) \rangle_{p(w|\theta)}$ found approximately by sampling, it would be an example of an EDA. The term was first used in the field of bioinformatics \cite{muehlenbein_paas}. EDAs have been widely developed by the evolutionary computing community \cite{lozanobook} and  applied to both discrete (for example \cite{pbil}) and continuous \cite{rudlof97} problems.

\section{Conclusion}

We have considered a variational optimization method which is applicable to problems whose properties normally prohibit gradient based algorithms. We have shown that the variational objective can be differentiable for non-differentiable and discontinuous objectives. The experimental results show the ability of the method to provide good approximate solutions to the lasso and soft margin support vector machine problems. It guarantees a bound on the solution, and can conserve convexity properties of the objective. The bound is interesting since the procedure to find a bound on the optimum value is straightforward, even in cases where a direct bound on the objective function itself is not apparent. Note that in our two example cases, computing the expectation over the parameters could be carried out analytically. In cases where the expectation can't be performed analytically, provided a distribution $p(x|\theta)$ is chosen for which samples can be drawn easily, then an approximate version of VO can be readily obtained numerically. Our description also makes a relation to alternative sampling based EDA procedures which can be viewed as special cases of VO and as such may motivate those methods as providing approximate bounds on the optima of the objective. 

In summary it is our view that the simplicity and generality of variational optimization lends itself to further study, with the potential for application to other interesting objectives.

\bibliographystyle{unsrt}
\bibliography{variationaloptimization}

\end{document}